\documentclass[sigconf]{acmart}

\usepackage{amsmath,amsfonts}
\usepackage{algorithmic}
\usepackage{graphicx}
\usepackage{textcomp}
\usepackage{todonotes}
\usepackage{xcolor}
\usepackage{url}
\usepackage{multicol}
\usepackage{array}
\usepackage{hyperref}
\usepackage{cleveref}

\usepackage{soul}
\usepackage{xcolor}

\newcommand{\sanr}{\texttt{SANR}}
\newcommand{\sanrl}{\texttt{SANRlite}}
\sethlcolor{yellow} 

\AtBeginDocument{%
  \providecommand\BibTeX{{%
    \normalfont B\kern-0.5em{\scshape i\kern-0.25em b}\kern-0.8em\TeX}}}

\setcopyright{none}
\copyrightyear{2024}
\acmYear{2024}
\acmDOI{XXXXXXX.XXXXXXX}

\acmConference[X]{Y}{Y}{2024}

\acmISBN{978-1-4503-XXXX-X/18/06}
\renewcommand\footnotetextcopyrightpermission[1]{}
\graphicspath{{./images/}}

\begin{document}

\title{Seventeenth-Century Spanish American Notary Records for Fine-Tuning Spanish Large Language Models}

\author{Shraboni Sarker}
\affiliation{%
  \institution{University of Missouri, Columbia}
  \country{USA}
}
\email{sscx3@umsystem.edu}

\author{Ahmad Tamim Hamad}
\affiliation{%
  \institution{University of Missouri, Columbia}
  \country{USA}
}
\email{ahzkc@missouri.edu}

\author{Hulayyil Alshammari}
\affiliation{%
  \institution{University of Missouri, Columbia}
  \country{USA}
}
\email{hamkx@missouri.edu}

\author{Viviana Grieco} 
\affiliation{%
 \institution{University of Missouri-Kansas City}
 \country{USA}
} 
\email{griecov@umkc.edu}

\author{Praveen Rao}
\affiliation{%
 \institution{University of Missouri, Columbia}
 \country{USA}
 }
\email{praveen.rao@missouri.edu}

\begin{abstract}
Large language models have gained tremendous popularity in domains such as ecommerce, finance, healthcare, and education. Fine-tuning is a common approach to customize an LLM on a domain-specific dataset for a desired downstream task. In this paper, we present a valuable resource for fine-tuning LLMs developed for the Spanish language to perform a variety of tasks such as classification, masked language modeling, clustering, and others. Our resource is a collection of handwritten notary records from the seventeenth century obtained from the National Archives of Argentina. This collection contains a combination of original images and transcribed text (and metadata) of 160+ pages that were handwritten by two notaries, namely, Estenban Agreda de Vergara and Nicolas de Valdivia y Brisuela nearly 400 years ago. Through empirical evaluation, we demonstrate that our collection can be used to fine-tune Spanish LLMs for tasks such as classification and masked language modeling, and can outperform pretrained Spanish models and ChatGPT-3.5/ChatGPT-4o. Our resource will be an invaluable resource for historical text analysis and is publicly available on GitHub.


\end{abstract}


\maketitle
\thispagestyle{empty}



\section{Introduction}

Large language models (LLMs) have gained tremendous popularity in domains such as ecommerce, finance, life sciences, healthcare, and education. However, these models are costly and time-consuming to train from scratch due to two reasons: First, they require large volumes of training data. Second, the training process demands advanced computing infrastructure such as 100's of graphics processing units (GPUs). Interestingly, Databricks recently released an open source LLM that cost \$10 million to create~\cite{DBRX}. Hence, fine-tuning on a domain-specific dataset is a common approach used to customize an LLM for a desired downstream task such as classification, sentiment analysis, question answering, and so on. Also, the growing availability of open source LLMs (e.g., Llama2~\cite{Llama2}, BERT~\cite{BERT}, DBRX~\cite{DBRX}) provides numerous opportunities for users to perform fine-tuning.

Historians and paleographers rely on historical documents to construct comprehensive narratives of a society's social, economic, cultural, and political developments. Historical documents require expert paleographers to read and transcribe these documents. Manual transcription is a time consuming process. For example, the notary records from $17^{th}$-century Spanish America, stored at the Archivo General de la Nación Argentina (National Archives)~\cite{archieve3}, offer valuable insights into the period. Typically, individuals proficient in Spanish require approximately one hour to carefully peruse a four-to-five-page notarized document, highlighting the detailed nature of the content~\cite{AmericanPoleographies14,wasserman2019escritura}. Hence, transcribing historical text is a challenging process especially when the availability of expert paleographers is limited.



Deep learning techniques~\cite{Book8} have shown tremendous promise in analyzing handwritten historical documents ranging from Arabic to Spanish languages~\cite{lombardi2020deep,sumac_alrasheed,MMAsia-23,IEEE-2019,DATeCH-14}. With the growing popularity and availability of LLMs, we envision new research efforts will harness the power of LLMs for historical text analysis and information retrieval in general. However, rich datasets are necessary to fine-tune LLMs for analyzing historical text.

Motivated by this reason, we present a unique data resource for fine-tuning LLMs for the Spanish language to enable tasks such as classification, mask language modeling, and clustering. Note that Spanish is the second most popular language in the world. Our resource was produced using a subset of the 17th-century Spanish American Notary Records (\sanr) with the assistance of two expert paleographers. Referred to hereinafter as \sanrl, the resource contains 160+ images of notary records written by two notaries, namely, Estenban Agreda de Vergara and Nicolas de Valdivia y Brisuela nearly 400 years ago. To demonstrate the value of \sanrl, we conducted an empirical evaluation of two tasks, namely, sentence classification and masked language modeling by fine-tuning existing pre-trained Spanish LLMs~\cite{BETO,MBERT}. We also compared the performance of the fine-tuned LLMs with ChatGPT models via prompting. For classification, fine-tuned LLMs significantly outperformed ChatGPT-based models in terms of F1 score and accuracy. For masked language modeling, fine-tuned LLMs outperformed both the pre-trained model and ChatGPT models. 

We believe \sanrl{} will foster new innovations in historical text analysis, natural language understanding, and information retrieval in general.

The rest of the paper is organized as follows: Section~\ref{sec-background} provides background on LLMs and fine-tuning for specific tasks. Section~\ref{sec-dataset} describes \sanrl{} including its metadata. Section~\ref{sec-experiments} presents the evaluation of fine-tuned LLMs using \sanrl{} and comparison with ChatGPT models. Finally, we conclude in Section~\ref{sec-conclusion}.
\section{Background}
\label{sec-background}

The introduction of the Transformer architecture~\cite{transfomers} and the underlying concept of \textit{self-attention} have revolutionized the field of natural language processing (NLP). In recent years, LLMs such as BERT~\cite{BERT}, which are based on the Transformer architecture and trained on massive text corpus, have become a powerful tool for text analysis~\cite{LLM-classification}. LLMs have the exceptional ability to be used for several downstream NLP tasks such as named entity recognition, classification, sentiment analysis, sentence generation, question answering, clustering, and so on. Chatbots such as ChatGPT-3.5/ChatGPT-4o~\cite{chatgpt35,chatgpt40} are LLMs trained on massive text corpus (e.g., Wikipedia, books) and can also be used for text analysis via prompting.  


As pre-training LLMs is expensive and time consuming, fine-tuning these models on specific datasets is commonly performed for certain tasks (e.g., masking, classification, named entity recognition). Thus, LLMs can adapt their generalized language understanding to the unique characteristics of the target domain. For instance, models like BETO~\cite{BETO} and M-BERT~\cite{MBERT} are fine-tuned BERT models on domain-specific data (e.g., Spanish corpus, Turkish corpus, Bengali corpus) and task-specific parameters achieving impressive results. In a recent study, researchers used Spanish radiological reports as the dataset for fine tuning M-BERT and obtained an F1 score of 0.7~\cite{ACL2020}. Nolazco-Flores et al.~\cite{IEEE2023} used BERT~\cite{BERT} and RoBERTa~\cite{Liu2019RoBERTaAR} pre-trained models to categorize the genre of Spanish books. After fine-tuning, they achieved 90\% accuracy in predicting categories such as "cooking books," "economics, business, and enterprise," and "juvenile books".

In recent years, deep learning has become popular for analyzing historical text. The Spanish American Notary Records from the National Archives of Argentina has been successfully used for character recognition~\cite{Character_recognition}, word recognition~\cite{sumac_alrasheed}, knowledge graph-based information retrieval system~\cite{KGSAR}, and few-shot learning for tackling different handwritings~\cite{MMAsia-23}. We believe LLMs fined-tuned on such datasets can be valuable for downstream processing tasks such as classification, masked language modeling, clustering, and others.

\section{Description of the Resource}
\label{sec-dataset}

\sanr{} from the National Archives of Argentina contains 220,000 images of notary records from the $17^{th}$ century. The documents in \sanr{} were written nearly 400 years ago and contain handwritten documents by different notaries. They cover a wide range of topics such as marriage deeds, business deeds, power of attorney, and so on. These images are organized into rollos (or rolls in Spanish) at the National Archives of Argentina. 

\begin{figure}[tbh]
\begin{center}
\includegraphics[width=3.3in]{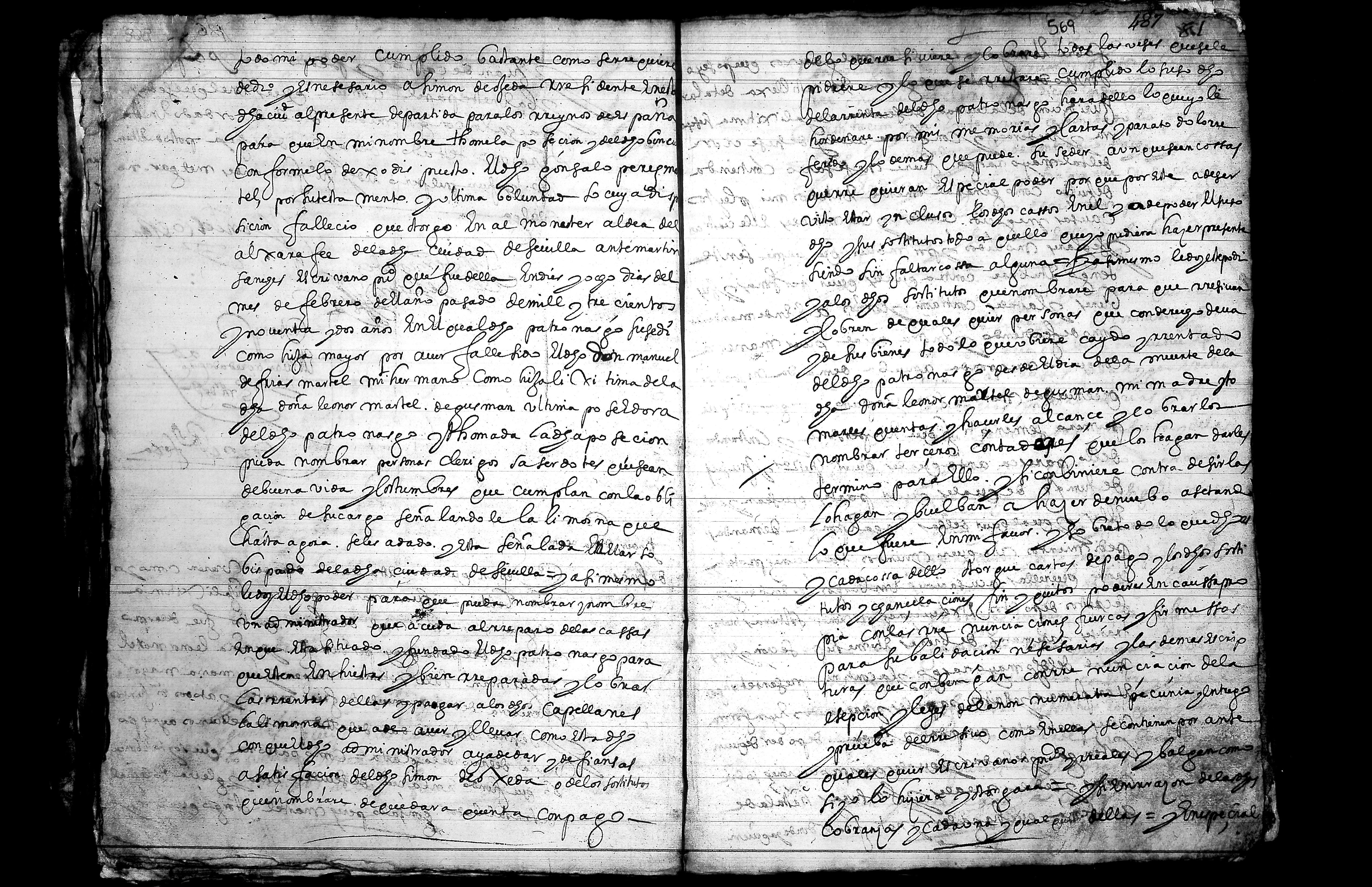}
\end{center}
\caption{An example image in \sanrl}
\label{fig-example-image}
\end{figure}

\begin{figure}[tbh]
  \centering
  \includegraphics[width=3.0in]{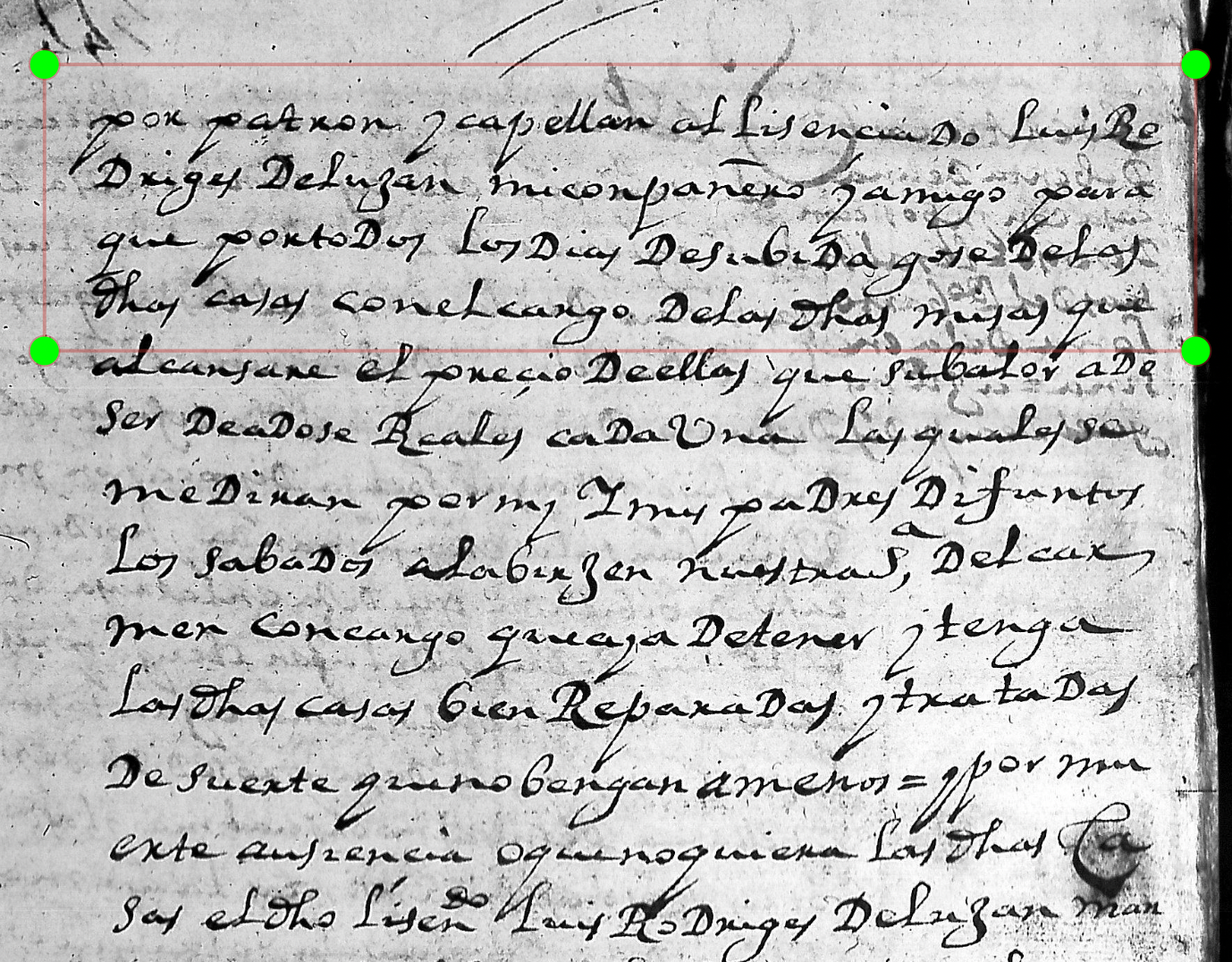}
  \caption{An example of a notary record in \sanrl{} and its annotation}
  \label{fig-notary-record}
\end{figure}

Our resource, \sanrl, is a subset of the original dataset and is available on GitHub~\cite{SANRlite}. To create \sanrl, we chose notary records written by two notaries, namely, Esteban Agreda de Vargara and Nicolas de Valdibia y Brizuela. Two Spanish paleography experts transcribed and annotated 81 double-page images from \textit{Rollo 38} and \textit{Rollo 40}. An example image is shown in Figure~\ref{fig-example-image}. An example of a portion of a notary record page and its annotation is shown in Figure~\ref{fig-notary-record}. Each page can have several sentences; a sentence can span across two pages. As the notary texts lacked sentence breaks, the paleography experts diligently separated the notary paragraphs into multiple sentences, ensuring that each sentence made sense semantically. In total, \sanrl{} had 162 pages containing 900+ sentences.

Each sentence (or a group of sentences) was assigned one or more class labels and extended class labels. Extended class labels provide fined-grained representation. For example, Figure~\ref{fig-JSON-metadata} shows the JSON metadata for a sample sentence. This sentence has two class labels, namely, "Parties" and "Scope". The extended class label for "Scope" is "Scope of deed" in this particular example. Other examples of extended class labels include "Header - power of attorney to represent in trials" or "Header - power of attorney to represent in trials and receive payments" for the class label "Header". In total, there are 33 class labels and 154 extended class labels that were assigned to the sentences. Most sentences had only one class label; however, some of the sentences were grouped together in a JSON record under \texttt{content}. 

To semantically enrich the JSON metadata, for each class label, we searched Wikidata~\cite{Wikidata}, a popular free and open knowledge base, to extract the uniform resource identifier (URI) for the class labels to precisely denote their meaning. The JSON metadata also includes the notary name, the year when the notary record was written, and the Rollo/image number. From the image number, the annotated images can be obtained via our GitHub repository~\cite{SANRlite}. Note that the images were annotated using LabelImg~\cite{LabelImg}. For each image, an XML document was generated by LabelImg that contained the coordinates of the annotation. 

We also generated sentence embeddings (of dimension 768) using masked language modeling. These embeddings can be used for training classifiers and performing clustering.

\begin{figure}[tbh]
\begin{small}
\begin{verbatim}
    {
        "id": "316",
        "year": 1658,
        "author_name": "Nicolas de Valdivia y Brisuela",
        "rollo_number": 40,
        "image_number": 182,
        "content": "por patron y capellan al
        lisençiando luis rodriges delujan mi
        companero y amigo para que portodos los
        dias desubida gose delas dhas casas.",
        "extended_class_label": [
            "Parties",
            "Scope of deed"
        ],
        "class_label": [
            "Parties",
            "Scope"
        ],
        "wikidata_concept": [
            "https://www.wikidata.org/wiki/Q60981922",
            "https://www.wikidata.org/wiki/Q472160"
        ]
    }
    \end{verbatim}
\end{small}
\vspace{-2ex}
\caption{JSON metadata for a sentence in a notary record}
\label{fig-JSON-metadata}
\end{figure}

\section{Experiments}
\label{sec-experiments}

To evaluate the benefit of \sanrl, we fine-tuned existing Spanish LLMs for classification and masked language modeling. We compared the performance of these fine-tuned models with ChatGPT for the aforementioned tasks.

\subsection{Implementation, Setup \& Evaluation Metrics}

We used the PyTorch Lightning code base for the multi-label classification task~\cite{MultilabelTextClassification}. We obtained code developed in Python from Hugging Face for masked language modeling~\cite{MLMCode}. All the experiments were conducted on a server machine with 2 Intel Xeon W-2245 8-core processors, 125 GB of RAM, 1 TB solid-state drive (SSD) storage, and 1 TB disk drive running Ubuntu 22.04.3 LTS. This machine had two NVIDIA RTX A4000 GPUs with 64 GB of GPU memory. 

For classification, we computed the standard metrics of F1 score, precision, recall, and accuracy. For masked language modeling, we computed F1 score, cosine similarity/exact match for word-level prediction, and BLEU score for sentence-level prediction~\cite{BLEU,cosine-similarity,ExactMatchMetric}.

\begin{figure*}[tbh]
\centering
\includegraphics[width=3.6in]{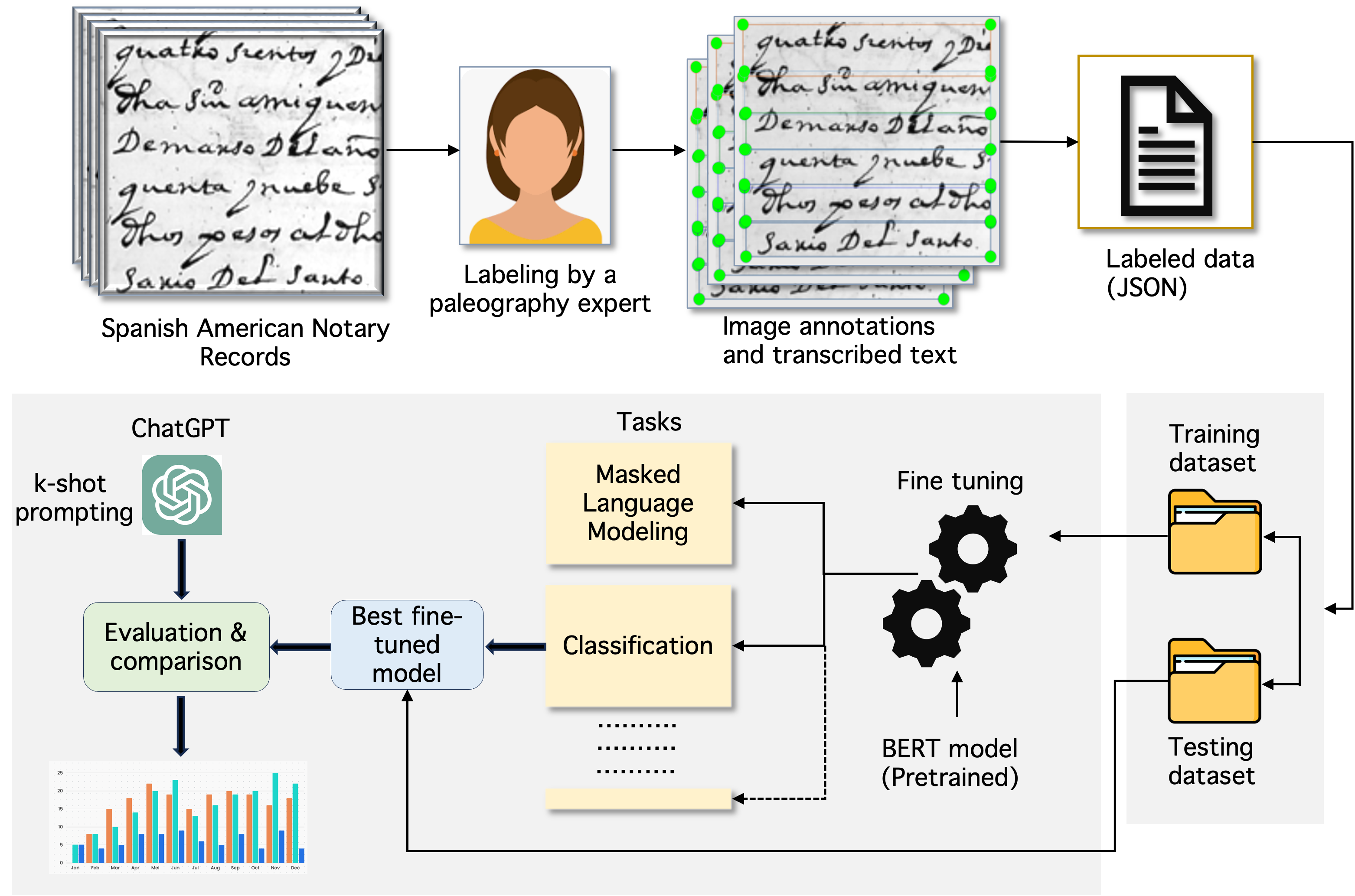}
\caption{Steps for constructing \sanrl{} and fine tuning BERT-based models for classification and mask language modeling}
\label{fig-overall-steps}
\end{figure*}



\subsection{Model Fine-Tuning}

We used BERT-based models~\cite{BERT} for fine tuning using \sanrl. These models had a hidden size of 768, 12 transformer blocks, and 12 self-attention heads. We chose BETO~\cite{BETO}, which is a pre-trained Spanish BERT model trained solely on a Spanish corpus using the Whole Word Masking technique. This model employed a vocabulary of approximately 31K BPE (Byte Pair Encoding) subwords constructed using \textit{SentencePiece} and underwent 2 million training steps. BETO demonstrated state-of-the-art performance across various NLP tasks in the Spanish language. We considered both cased and uncased versions of this model~\cite{BETO}. We also chose a BERT-based multilingual model called M-BERT~\cite{MBERT}, which was pre-trained on the top 104 languages with the largest Wikipedia corpus using masked language modeling objectives. We considered both cased and uncased versions of this model. 

\sanrl{} contained 952 sentences obtained from 81 double-page images. We split \sanrl{} into 80\% for training, 10\% for validation, and 10\% for testing. For classification and masked language modeling, we also trained the tokenizer. Figure~\ref{fig-overall-steps} shows the overall steps starting from the original images, to annotating the sentences in these images, to generating metadata, and finally, to fine tuning the BERT-based models for evaluation and comparison.

\begin{table*}[tbh]
  \caption{Classification Performance: Fine-tuned Spanish LLMs and ChatGPT (best results in bold)}
  \label{tab:performance_metrics}
  \centering
  \begin{tabular}{|c|c|c|c|c|c|c|c|c|c|c|}
  \hline
    \multicolumn{1}{|c}{} &
    \multicolumn{4}{|c}{\textbf{Our Approach (Fine-Tuning)}} & \multicolumn{3}{|c|}{\textbf{ChatGPT-3.5}} & 
    \multicolumn{3}{|c|}{\textbf{ChatGPT-4o}} \\
    \cline{2-11}
    \multicolumn{1}{|c}{\textbf{Metric}} &
    \multicolumn{1}{|c|}{BETO} &
    \multicolumn{1}{|c|}{BETO} &
    \multicolumn{1}{|c|}{M-BERT} &
    \multicolumn{1}{|c|}{M-BERT} &
    \multicolumn{1}{|c|}{Zero-Shot} &
    \multicolumn{1}{|c|}{Four-Shot} &
    \multicolumn{1}{|c|}{Five-Shot} &
    \multicolumn{1}{|c|}{Zero-Shot} &
    \multicolumn{1}{|c|}{Four-Shot} &
    \multicolumn{1}{|c|}{Five-Shot} \\
    \multicolumn{1}{|c}{} &
    \multicolumn{1}{|c|}{(cased)} &
    \multicolumn{1}{|c|}{(uncased)} &
    \multicolumn{1}{|c|}{(cased)} &
    \multicolumn{1}{|c|}{(uncased)} &
    \multicolumn{1}{|c|}{} &
    \multicolumn{1}{|c|}{} &
    \multicolumn{1}{|c|}{} &
    \multicolumn{1}{|c|}{} &
    \multicolumn{1}{|c|}{} &
    \multicolumn{1}{|c|}{} \\
    \cline{2-11}
    \hline
    \hline
    F1 score & 0.688 & 0.688 & \textbf{0.713} & 0.660 & 0.164 & 0.124 & 0.289 & 0.185 & 0.274 & 0.110 \\
    \hline
    Accuracy & 63.5\% & 62.5\% & \textbf{64.5\%} & 59.3\% & 15.1\% & 10.2\% & 22.7\% & 14.7\% & 25.0\% & 10.2\% \\
    \hline
  \end{tabular}
\end{table*}




\subsection{Multi-Label Classification: Fine-Tuned LLMs Versus Pre-Trained LLMs}

We used the BERT-based models for multi-label classification of sentences. We fine-tuned 4 different pre-trained BERT models: (a) BETO: Spanish BERT (cased), (b) BETO: Spanish BERT (uncased), (c) Multilingual BERT (M-BERT) (cased), and (d) M-BERT (uncased). We used different hyperparameters (e.g., batch size, learning rate, optimizer) during fine-tuning and chose the best models for evaluation on the test set. To gained deeper insights, we compared the performance of fine-tuned models with ChatGPT-3.5~\cite{chatgpt35} and ChatGPT-4o~\cite{chatgpt40}. We used zero-shot, four-shot, and five-shot prompting for ChatGPT models.

The classification results are shown in Table~\ref{tab:performance_metrics}. The fine-tuned M-BERT (cased) outperformed all the other models achieving the best F1 score of 0.713 and accuracy of 64.5\%. In general fined-tuned BERT-based models using \sanrl{} performed significantly better than ChatGPT 3.5 and ChatGPT-4o via few-shot prompting. 



\begin{table*}[tbh]
  \caption{Masked language modeling: Fine-tuned BETO versus pre-trained BETO (best results in bold)}
  \label{tab:mlm-performance}
  \begin{tabular}{|c|c|c|c|c|c|c|c|c|}
  \hline
  \multicolumn{1}{|c|}{\textbf{Model}} &
  \multicolumn{1}{|c|}{\textbf{Batch Size}} & 
  \multicolumn{1}{|c|}{\textbf{Cosine Similarity}} &
  \multicolumn{1}{|c|}{\textbf{Sentence BLEU}} &
  \multicolumn{1}{|c|}{\textbf{Corpus BLEU}} &
  \multicolumn{1}{|c|}{\textbf{Exact Match}} &
   \multicolumn{1}{|c|}{\textbf{F1 Score}} \\
  \hline
  \hline
    \texttt{BETO} (fine-tuned) & 8 & 0.818 & 0.700 & 0.715 & 0.202 & 0.129\\
    \hline
    \texttt{BETO} (fine-tuned) & 16 & 0.878 & 0.720 & 0.725 & 0.418 & \textbf{0.243}\\
    \hline
    \texttt{BETO} (fine-tuned) & 32 & \textbf{0.878} & \textbf{0.746} & \textbf{0.752} & \textbf{0.432} & 0.240\\
    \hline
    \texttt{BETO} (pre-trained) & -- & 0.749 & 0.702 & 0.714 & 0.004 & 0.006\\
    \hline
  \end{tabular}
\end{table*}

\subsection{Masked Language Modeling: Fined-Tuned LLMs Versus Pre-Trained LLMs}

We used BETO (cased) for evaluating the performance of masked language modeling using \sanrl. We used a masking rate to 10\% for training and testing. We used different hyperparameters (e.g., batch size, chunk size, learning rate) during fine-tuning and chose the best models for evaluation on the test set. To compute the evaluation score for a model, we obtained the top 3 model predictions for a sentence in the test set and calculated the average score across these predictions. Finally, we computed the average score across the test set and report a single score for a fine-tuned model. 



The performance results are shown in Table~\ref{tab:mlm-performance}. Most of the fine-tuned models outperformed the pre-trained BETO model. When using a batch size of 32, we observed that the fine-tuned BETO model achieved the highest cosine similarity in word-level predictions and sentence/corpus BLEU score in sentence-level predictions. Notably, we achieved the highest exact match score for a batch size of 32 in fine-tuned BETO, and for a batch size of 16, we obtained the best F1 score. Once again, our fine-tuned models using \sanrl{} performed better than pre-trained models for masked language modeling.



\section{Conclusion}
\label{sec-conclusion}

In this resource paper, we presented \sanrl, which was obtained from the National Archives of Argentina. It serves as a valuable resource for fine-tuning Spanish LLMs for tasks such as classification and masked language modeling. The resource contains images and transcribed text (of 160+ pages) along with necessary metadata (in JSON) such as Wikidata concepts. It also contains the fine-tuned LLMs and sentence embeddings that can be used for clustering and other tasks. We demonstrated that fined-tuned Spanish LLMs on our collection can outperform ChatGPT-3.5/ChatGPT-4o for sentence classification and masked language modeling. We believe our resource will be of immense value to the historical text analysis community to foster new advances in natural language understanding, handwriting analysis, and information retrieval in general. In the future, we would like to explore how multimodal embeddings can be generated (e.g., using CLIP~\cite{CLIP2021}) by learning on the sentences and image annotations.


\begin{acks}
This work was supported by the National Endowments for the Humanities Grant No. HAA-287903-22. We would like to thank Dr. Martin Wasserman for his assistance in transcribing certain notary records. The third author (H. A.) would like to acknowledge the support of Saudi Arabian Cultural Mission (SACM).
\end{acks}

\bibliographystyle{ACM-Reference-Format}
\bibliography{reference}

\end{document}